\documentclass[10pt,twocolumn,letterpaper]{article}

\usepackage{icb}
\usepackage{times}
\usepackage{epsfig}
\usepackage{graphicx}
\usepackage{amsmath}
\usepackage{amssymb}
\usepackage{gensymb}

\usepackage{url}

\icbfinalcopy 

\usepackage{fancyhdr}
\begin{document}

\title{Post-mortem Human Iris Recognition}

\author{Mateusz Trokielewicz$^{\dag,\ddag}$, Adam Czajka$^{\ddag,\dag}$\\
$^{\dag}$Research and Academic Computer Network\\
Wawozowa 18, 02-796 Warsaw, Poland\\
$^{\ddag}$Institute of Control and Computation Engineering\\
Warsaw University of Technology\\
Nowowiejska 15/19, 00-665 Warsaw, Poland\\
{\tt\small {mateusz.trokielewicz,adam.czajka}@nask.pl}
\and
Piotr Maciejewicz$^{\star}$\\
$^{\star}$Department of Ophthalmology\\
Medical University of Warsaw\\
Lindleya 4, 02-005 Warsaw, Poland\\
{\tt\small piotr.maciejewicz@wum.edu.pl}
}

\maketitle
\thispagestyle{empty}

\begin{abstract}
This paper presents a unique analysis of post-mortem human iris recognition. Post-mortem human iris images were collected at the university mortuary in three sessions separated by approximately 11 hours, with the first session organized from 5 to 7 hours after demise. Analysis performed for four independent iris recognition methods shows that the common claim of the iris being useless for biometric identification soon after death is not entirely true. Since the pupil has a constant and neutral dilation after death (the so called ``cadaveric position''), this makes the iris pattern perfectly visible from the standpoint of dilation. We found that more than 90\% of irises are still correctly recognized when captured a few hours after death, and that serious iris deterioration begins approximately 22 hours later, since the recognition rate drops to a range of 13.3-73.3\% (depending on the method used) when the cornea starts to be cloudy. There were only two failures to enroll (out of 104 images) observed for only a single method (out of four employed in this study). These findings show that the dynamics of post-mortem changes to the iris that are important for biometric identification are much more moderate than previously believed. To the best of our knowledge, this paper presents the first experimental study of how iris recognition works after death, and we hope that these preliminary findings will stimulate further research in this area.
\end{abstract}

\section{Introduction}
Biometric recognition of living individuals with the use of their iris patterns has been thoroughly studied and has received considerable attention in the biometric community. However, the aspect of performing post-mortem iris recognition remains fairly unstudied. This maybe disagreeable aspect is, however, crucial from the forensics point of view. The most important motivation for this study was to know if one could use iris to verify the identity of a dead person. This addresses also a social anxiousness suggesting that someone may steal iris images of a death person and use them in presentation attacks.

A common belief that a dead iris is completely unusable originates probably from Daugman's statement that ``soon after death, the pupil dilates considerably, and the cornea becomes cloudy'' \cite{DaugmanPostMortem}. This opinion was then used throughout the years in both scientific papers and commercial descriptions. For instance, Szczepanski {\it et al.} \cite{SaeedPostMortem} claim that ``the iris (...) decays only a few minutes after death'', yet present no experimental evidence for this claim. When analyzing commercial iris sensors and solutions, we may learn that ``the notion of stealing someone's iris after death is scientifically impossible. The iris is a muscle; it completely relaxes after death and results in a fully dilated pupil with no visible iris at all. A dead person simply does not have a usable iris!'' \cite{IrisGuardPostMortem}, or that ``after death, a person's iris features will vanish along with pupil's dilation'' \cite{IriTechPostMortem}. 

This paper shows that the above claims are only partially true. We had a rare possibility to organize three iris image acquisition sessions, separated by 11 hours, at the university mortuary, with the first session organized approximately 5 hours after death. A commercial iris recognition camera operating in NIR light, as well as a color camera were used. The data acquisition process is detailed in Section \ref{sec:DataCollection}. We found that pupil does not excessively dilate and has rather a neutral size, making the iris perfectly visible. 

Results presented in Section \ref{sec:Results} show that the iris indeed degrades after death, but that, under mortuary conditions, the effect on iris recognition takes hours rather than minutes. One of the iris matchers is still able to correctly recognize more than 70\% of irises 27 hours after death, and all four matchers employed in this work correctly match more than 90\% of iris pairs for 5 to 7 hours after death. In order to make this paper complete in terms of both engineering and medical aspects, a medical commentary on post-mortem iris changes is given in Section \ref{sec:Medical}.

\section{Post-mortem changes in the eye}
\label{sec:Medical}
After death, a sequence of changes naturally occurs in a human body. Those changes begin at the molecular level and sequentially progress to microscopic and gross morphology. Various body fluids show chemical changes after death, each with its own time factor. For instance, the levels of potassium concentration are increasing in the vitreous humor in the posterior chamber of the eye. Occasionally, this can prove useful for forensic pathologists to ascertain time since death in cases of unwitnessed demise. Desiccation (\emph{desiccatio post mortem}) with stiffening of the body (\emph{rigor mortis}), lividity (\emph{livors mortis}), body cooling (\emph{algor mortis}) and paleness (\emph{pallor mortis}) are five natural occurrences within the human body after death. The cornea and conjunctiva, being superficial tissues, are fairly prone to drying.

\subsection{Corneal transparency loss and tissue decay}
The cornea is an avascular transparent convex refractile -- a colorless structure that overlies the anterior chamber of the eye. Its transparency is essential for useful vision and for good visualization of the ocular structures located behind it, in the anterior chamber or in the posterior segment of the eye. Corneal transparency depends on well controlled hydration maintained by pre-corneal tear film. When lacrimal glands secretion and blinking cease, anoxia, dehydration and acidosis lead to autolysis (self-destruction) of the corneal cells and damage of the corneal surface, causing opacification that intensifies with time elapsed since death.

Apart from the opacity of the cornea there is escalating wrinkling of the corneal surface, flaccidity of the eyeball and sinking of the eyeball into the orbital cavity as time progresses. Usually, turbidity of the cornea will appear from a few hours to 24 hours after demise. It depends on the degree of eyelid closure, environmental temperature, humidity, and air movement. The ambient temperature has significant influence on the protein degradation and survival of corneal endothelial cells. Many other external and intrinsic factors may affect decomposition processes.

\subsection{Changes to the sclera}
\emph{Tache noire} is another example of changes that can be observed in a cadaver eye. If the eyes are not closed, the areas of the sclera exposed to air dry out. It results in a first yellowish, then brownish horizontal band across the eye ball -- a discoloration zone called \emph{tache noire}, or ``black line'', usually seen within 8 hours after death depending on environmental conditions. It is not visible over the cornea, though. 
 
 \subsection{Changes to the iris}
After death, pupils are usually mid-dilated and fixed in what is called a ``cadaveric position'', however, their shape and diameter may depend on previous medical history. For the first few hours after somatic death the iris can respond to local chemical stimulation, but does not react to light.

\section{Related work}
\label{sec:Related}

To the best of our knowledge, there are no scientific papers studying post-mortem iris recognition in the biometric sense (\ie, analyzing a performance of automatic iris recognition methods and estimating the feasibility to perform iris recognition in humans after death). In his presentation, Ross \cite{RossPostMortem} mentions the CITeR project `Post Mortem Ocular Biometric Analysis', however, the project outcomes are summarized in a few general conclusions. He states that ``the pupillary margin became indistinguishable in certain eye images, which made identifying the boundary of the pupillary margin difficult'' and that ``in certain cases iris tissues were difficult to differentiate from adjacent scleral tissues, making limbal boundary indistinct''. However, the number of samples analyzed, the time elapsed between death and the analysis, and the ambient conditions present in this study remain unknown.

Recently, Saripalle \emph{et al.} \cite{PostMortemPigs} have provided an excellent account on post-mortem iris recognition performed with the eyes of a domestic pig. Researchers managed to show that after death and removal of the eye, it slowly degrades and this degradation is accompanied by a continuing increase in genuine dissimilarity scores (fractional Hamming distance, since an algorithm based on Daugman's idea was used for this research). In most cases, the {\it ex-vivo} eye (\ie, eye removed from the body) lost its ability to perform as a biometric identifier approximately 6-8 hours after death. However, there are several differences between pig and human eyes, including shape and size of the iris. Also, as authors themselves notice, the fact that verification was performed using {\it ex-vivo} eyes could have affected the rate of corneal opacification progress, making corneal tissue deterioration faster. 

Despite the lack of papers studying post-mortem iris recognition in humans, there are several studies dedicated to a somewhat related field of iris biometrics: iris recognition in the presence of ocular pathologies. As the inevitable drying and the following degradation of the corneal tissue seems to be potentially the most prevalent source of trouble in post-mortem iris biometrics, some observations are worth mentioning here. Aslam \emph{et al.} \cite{Aslam}, while studying the effects of eye pathologies on the iris recognition performance, noticed that near-infrared (NIR) illumination seems to better penetrate corneal defects, such as haze or opacities, therefore these defects do not seem to affect recognition accuracy significantly. Differences between visible and NIR images of eyes affected by pathologies related to the cornea were also reported by Trokielewicz \etal\cite{TrokielewiczBTAS2015, TrokielewiczCYBCONF2015} in their studies dedicated to eye pathologies and their impact on the reliability of iris recognition. Those differences are strongly in favor of NIR-illuminated photographs, as even irises heavily occluded by corneal defects are said to be imaged effortlessly in near infrared.

\section{Database of iris images}
\label{sec:DataCollection}
\subsection{Data collection protocol}
For the purpose of this study, a new database of iris images had to be constructed. It consists of images representing the eye region of six recently deceased persons (hence 12 distinct eyes), including the iris. Images have been acquired using a dedicated iris recognition camera operating in near infrared (NIR) light -- the IriShield M2120U, which provides images compliant with the ISO/IEC standards regarding biometric data quality \cite{ISO,ISO2}. In addition to that, the eyes have also been photographed using a color camera -- Olympus TG-3, for easier visual assessment of the post-mortem damage to the eye tissues that could prevent iris biometrics from working reliably.

Images were captured in two or three sessions, depending on the cadaver availability. The first session was conducted approximately 5-7 hours after death, and sessions two and three were carried out approximately 11-15 and 22 hours later, respectively. During each session, 2 to 7 images of each eye were taken using each camera. Table \ref{table:database} shows the time elapsed since death in each image acquisition session and each subject. The temperature in the hospital mortuary where data collection was performed was approximately 6\degree  Celsius (42.8\degree  Fahrenheit).

\begin{table}[!ht]
\renewcommand{\arraystretch}{1.1}
\caption{Approximate time elapsed since death (t) for each acquisition session and for each subject.}
\label{table:database}
\centering\footnotesize
\begin{tabular}[t]{|c|c|c|c|c|}
\hline
\textbf{Subject ID} & \textbf{Session 1} & \textbf{Session 2} & \textbf{Session 3}\\
\hline
\hline
\textbf{0002} & {t + 5 hours} & {t + 16 hours} & {t + 27 hours}  \\
\hline
\textbf{0003} & {t + 5 hours} & {t + 16 hours} & {t + 27 hours} 	 \\
\hline
\textbf{0004} & {t + 5-7 hours} & {t + 19.5-21.5 hours}   & n/a\\
\hline
\textbf{0005} &{t + 5-7 hours} & {t + 19.5-21.5 hours}   & n/a\\
\hline
\textbf{0006} & {t + 5-7 hours} & {t + 16-18 hours} & n/a \\
\hline
\textbf{0007} & {t + 5-7 hours} & {t + 16-18 hours} & n/a \\
\hline
\end{tabular}
\end{table}

\subsection{Visual inspection of the samples}
\begin{figure}[!t]
\centering
\includegraphics[width=0.49\textwidth]{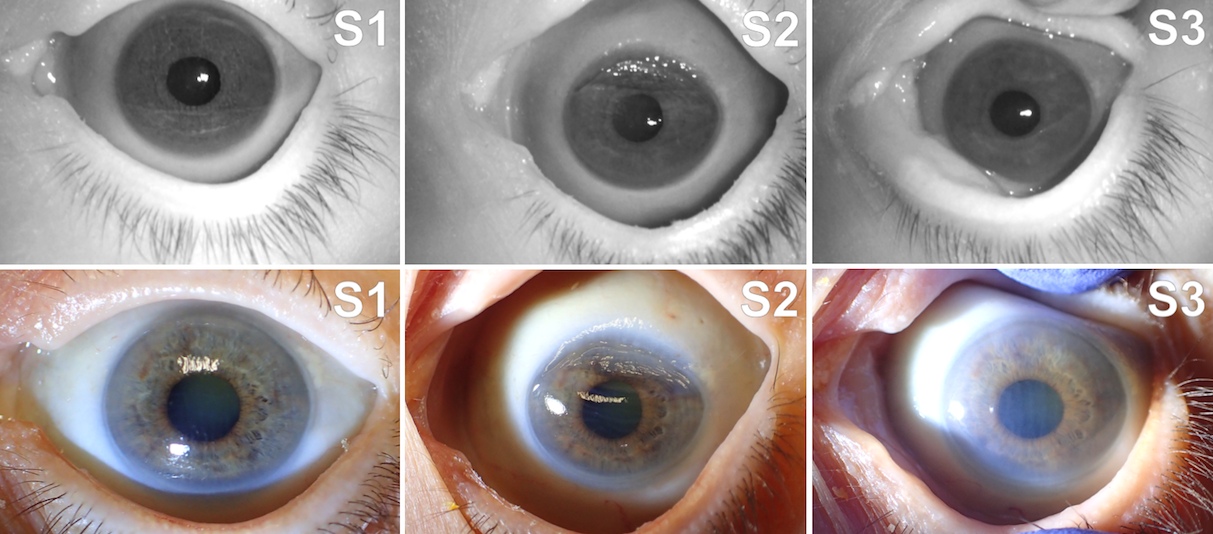}
\caption{Sample images from three acquisition sessions (Session 1, 2 and 3 - S1, S2 and S3, respectively) from both the NIR and visible light cameras. Significant changes to the eye can be spotted in S2 images, which later disappear in S3 images.}
\label{fig:samples_with_changes}
\end{figure}

\begin{figure}[!t]
\centering
\includegraphics[width=0.49\textwidth]{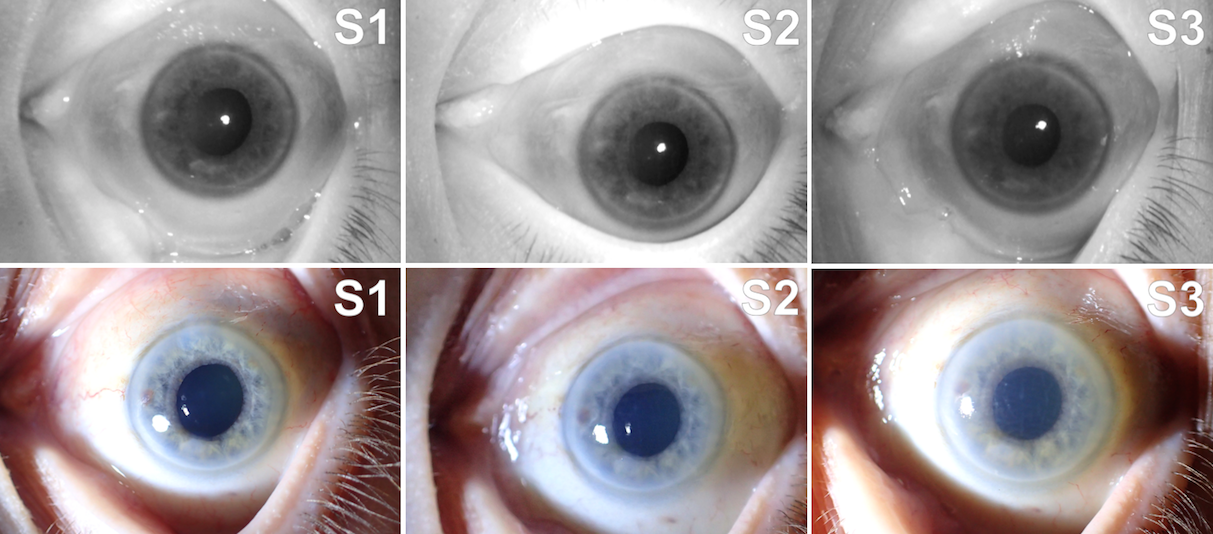}
\caption{Same as in Fig. \ref{fig:samples_with_changes}, but no changes to the eye, apart from increasing corneal opacity, can be seen in this case.}
\label{fig:samples_without_changes}
\end{figure}

Figures \ref{fig:samples_with_changes} and \ref{fig:samples_without_changes} show sample iris images obtained in each session, both NIR and visible light illuminated. Notably, significant changes regarding increasing corneal opacity can be seen in both cases. However, NIR images seem to show certain resilience against this phenomenon, as iris patterns are well visible throughout all three sessions (especially in Fig. \ref{fig:samples_without_changes}). Most interesting, however, are the changes in the iris pattern observed in images in Fig. \ref{fig:samples_with_changes}. In Session 2 images, a pattern distortion can be observed, which later disappears in Session 3 images, revealing an occluded, but otherwise unaffected iris pattern. This is attributed to the intentional compression applied to the eyeball using fingers, which temporarily counter-effected the loss of eyeball pressure.

\section{Experimental methodology}

\subsection{Genuine comparisons}

To explore the changes occurring in the iris after death, and assess whether iris recognition is feasible despite those changes, the following experiments have been carried out. All possible genuine comparisons between samples have been performed in the following three scenarios:
\begin{itemize}
	\item intra-session comparisons of the Session 1 images (referred to as \textbf{S1 vs S1} later on),
	\item inter-session comparisons of Session 2 images against Session 1 images (\textbf{S2 vs S1}),
	\item inter-session comparisons of Session 3 images against Session 1 images (when available, \textbf{S3 vs S1}).
\end{itemize}

This is also to find out whether there is a downward trend in recognition accuracy as time after death elapses. 

Table \ref{table:comparisons} shows the numbers of genuine comparisons performed in each case. Different numbers for the MIRLIN matcher result from two errors in template extraction by the MIRLIN method, hence providing a smaller number of templates to be matched.

\begin{table}[!ht]
\renewcommand{\arraystretch}{1.1}
\caption{Genuine comparison count for each method and each experimental scenario.}
\label{table:comparisons}
\centering\footnotesize
\begin{tabular}[t]{|c|c|c|c|c|}
\hline
\textbf{Method / Scenario} & \textbf{S1 vs S1} & \textbf{S2 vs S1} & \textbf{S3 vs S1} \\
\hline
\hline
\textbf{MIRLIN} & 70 & 174 & 26 \\
\hline
\textbf{VeriEye} & & & \\
\textbf{IriCore} & 72 & 178 & 30 \\
\textbf{OSIRIS} & & & \\
\hline
\end{tabular}
\end{table}

\subsection{Iris recognition methods}
For the purpose of generating comparison scores between iris images, four different, well-recognized iris recognition algorithms have been employed, including three commercial matchers and one open-source solution. This section provides a brief description of each of those methods. 

\textbf{VeriEye} \cite{VeriEye} utilizes an unpublished encoding methodology and iris localization that is said to employ a non-circular approximation of the iris and pupil boundaries. Comparison scores yielded by this method represent a similarity metric, utilizing values from 0 (for different irises) to some unknown, high values, typically exceeding a few hundred for same irises.

\textbf{MIRLIN} \cite{Monro2007}\cite{MIRLIN}, a method that has been available on the market for commercial applications, incorporates a Discrete Cosine Transform (DCT) calculated for the overlapping angular patches of the iris image. This results in a binary code of the iris pattern, which is then compared using XOR operation. Fractional Hamming distance is used as a comparison score, with values close to zero being expected for same-eye images, and values around 0.5 for different-eye ones. 

The third method, \textbf{OSIRIS} \cite{OSIRIS}, is an academically developed open-source implementation of the original Daugman's idea. Iris binary codes are calculated based on the complex plane phase quantization of the outcomes of Gabor-based filtering. This method, however, differs from Daugman's concept by employing active contours for image segmentation. Comparison scores are presented in the form of a fractional Hamming distance, similarly to the MIRLIN method. 

Finally, the \textbf{IriCore} \cite{IriCore} algorithm is a commercially available product, whose implementation details have not been revealed by the vendor. Comparison scores for same-eye images are expected to return values close to zero, while scores for different-eye images typically range from 1.1 to 2.0.

\section{Results}
\label{sec:Results}

\subsection{Template generation}

All methods seem to perform very well in terms of iris template generation. IriCore, VeriEye and OSIRIS were able to calculate templates for 100\% of samples collected in all three sessions. MIRLIN could not extract a template only for 2 images out of 104 (one acquired in the first session and one acquired in the third session), hence obtaining FTE = 1.9\%. (FTE, failure-to-enroll, is the proportion of template generation transactions returning an error to the overall number of enrollment transactions.) This suggests that the quality of biometric samples acquired in all three sessions was highly acceptable for iris recognition methods employed in this study.

\subsection{Recognition performance}

\begin{figure}[!htb]
\centering
\includegraphics[width=0.46\textwidth]{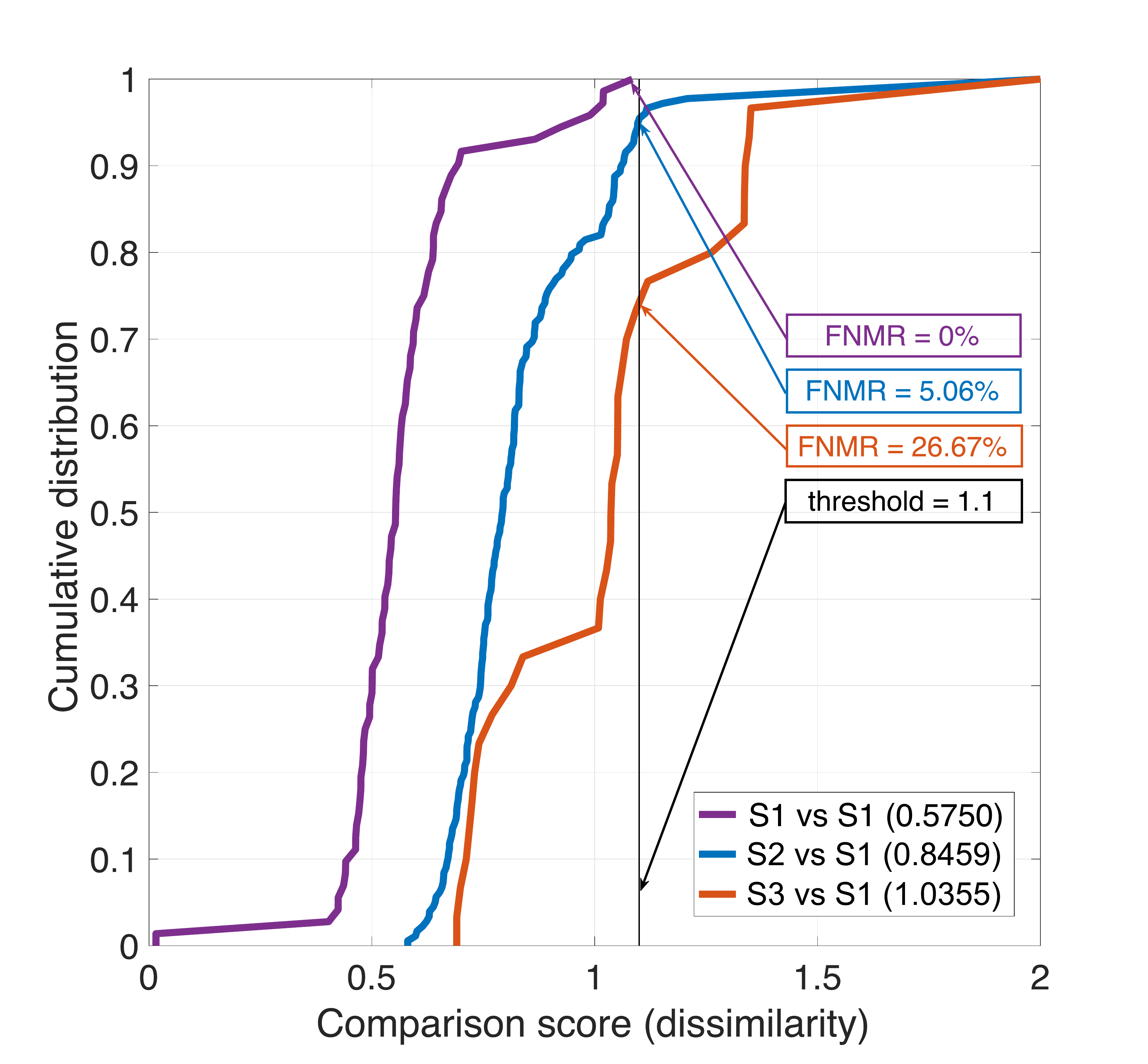}\hfill
\caption{Cumulative distribution functions of genuine comparison scores for the \textbf{IriCore} method. Intra-session results (Session 1 vs Session 1 samples) are shown in violet and inter-session results are shown in blue (Session 2 vs Session 1 samples) and red (Session 3 vs Session 1 samples). Sample FNMR is shown for the default acceptance threshold. Mean comparison scores obtained in three different experiments are also shown in brackets.}
\label{fig:CDF:IC}
\end{figure}

\begin{figure}[!htb]
\centering
\includegraphics[width=0.46\textwidth]{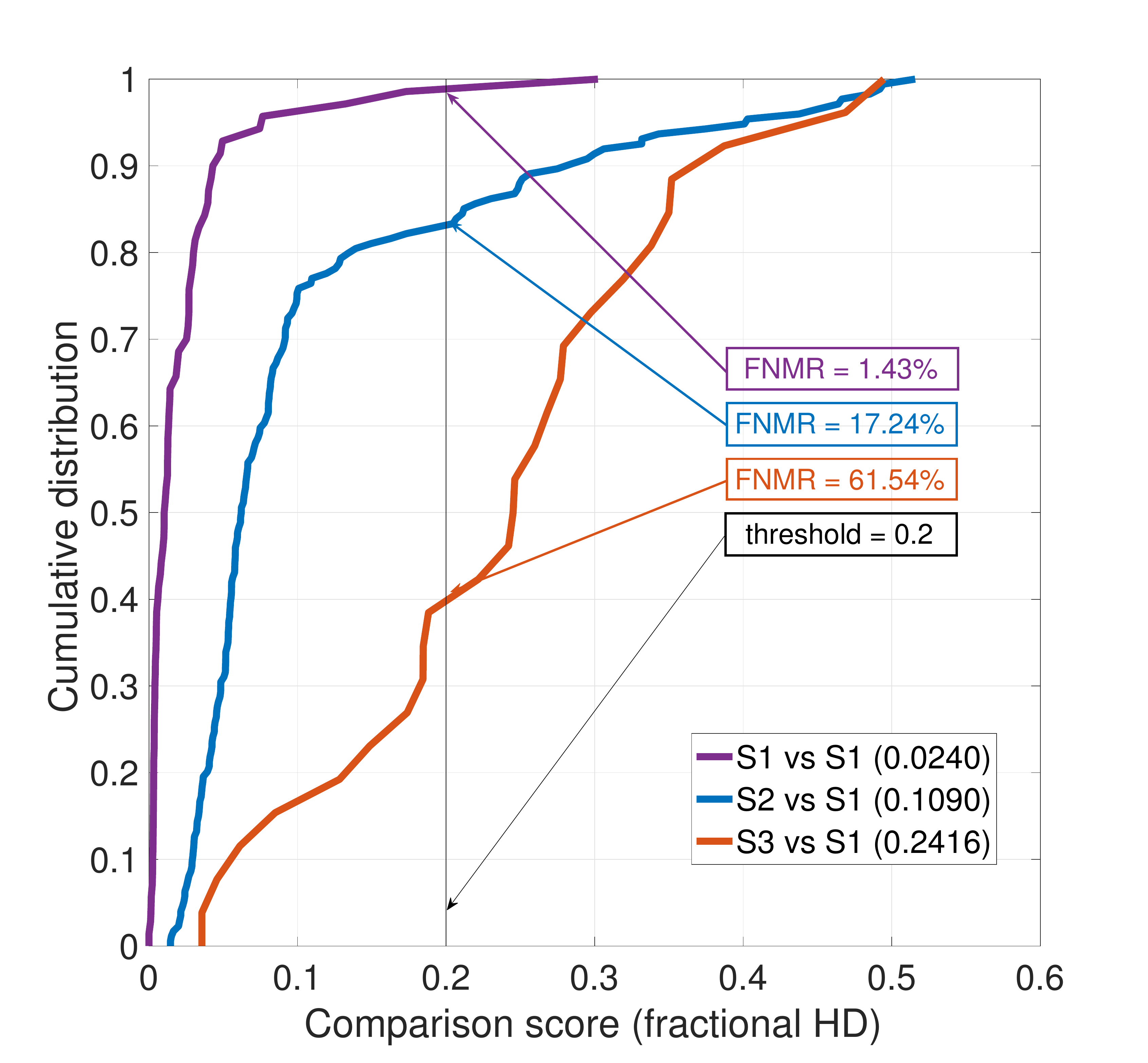}\hfill
\caption{Same as in Fig. \ref{fig:CDF:IC}, except for the \textbf{MIRLIN} method.}
\label{fig:CDF:ML}
\end{figure}

\begin{figure}[!htb]
\centering
\includegraphics[width=0.46\textwidth]{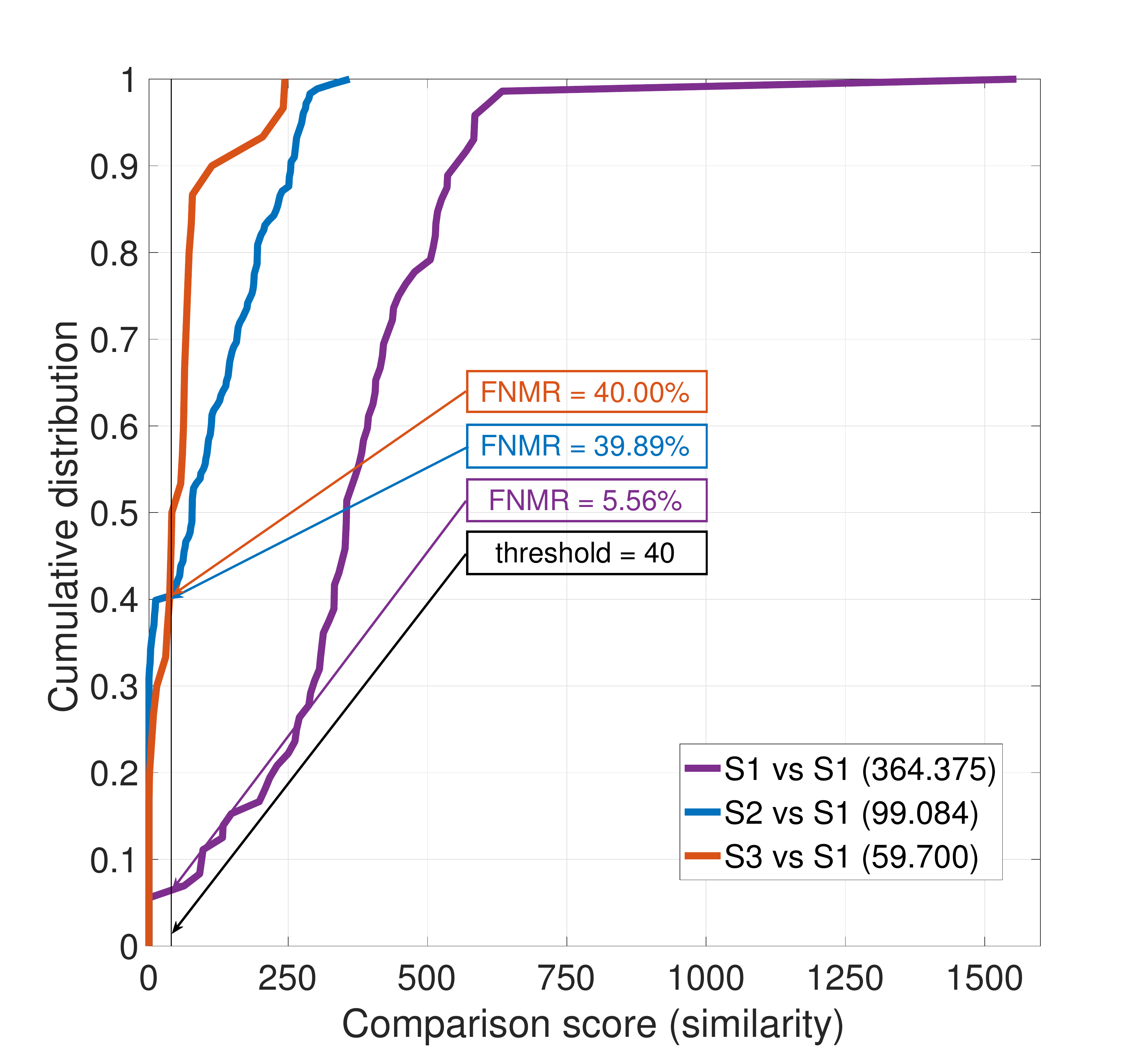}\hfill
\caption{Same as in Fig. \ref{fig:CDF:IC}, except for the \textbf{VeriEye} method.}
\label{fig:CDF:VE}
\end{figure}

\begin{figure}[!htb]
\centering
\includegraphics[width=0.46\textwidth]{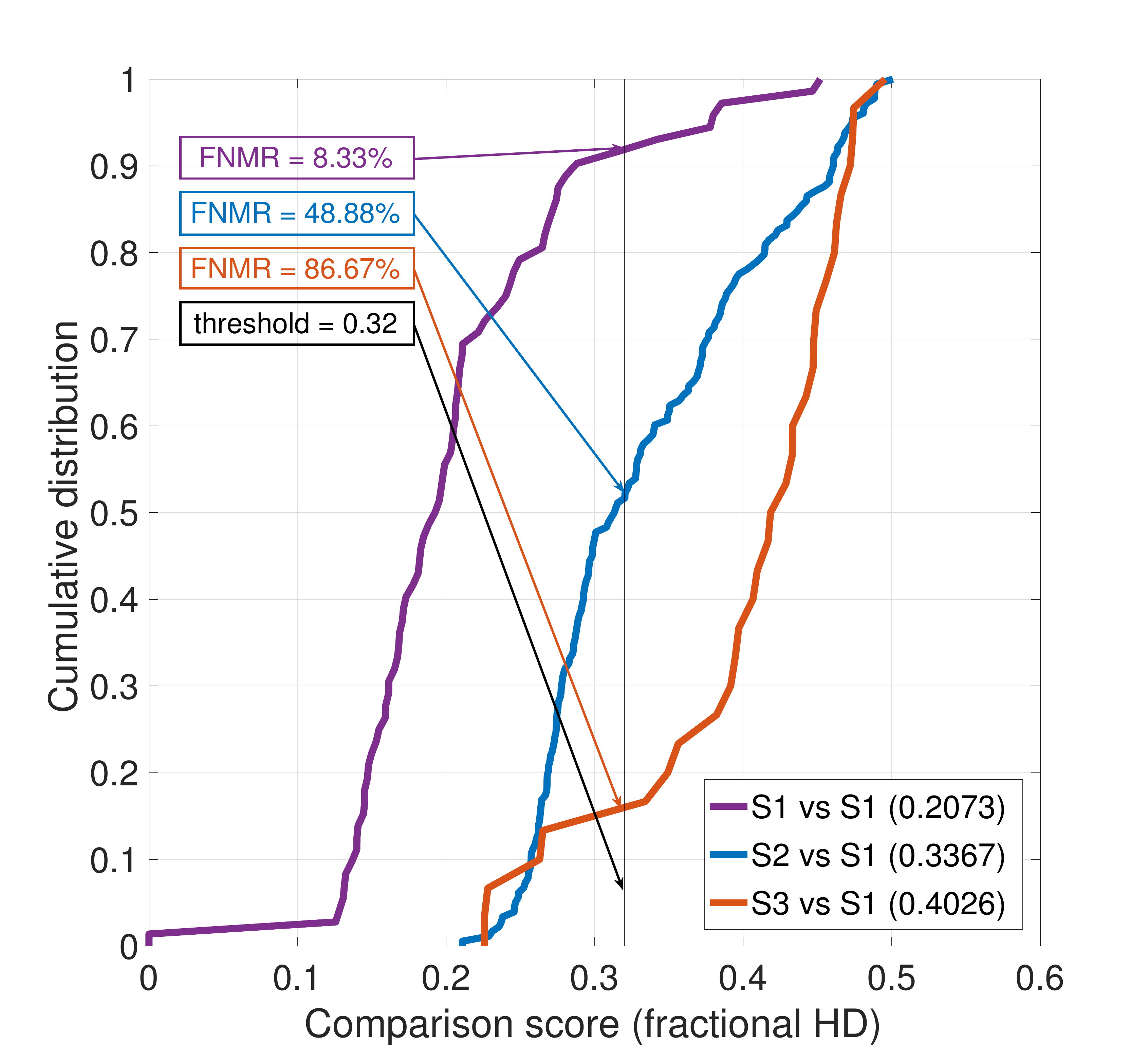}\hfill
\caption{Same as in Fig. \ref{fig:CDF:IC}, except for the \textbf{OSIRIS} method.}
\label{fig:CDF:OS}
\end{figure}

There are a few interesting findings when analyzing images and genuine comparison scores generated in this study. Figures \ref{fig:CDF:IC} through \ref{fig:CDF:OS} present sample cumulative distribution functions calculated for all three experiments (scenarios) and all four iris recognition methods.

The first observation, positively correlating with medical knowledge in the field, is that the pupil has rather a neutral size, making the iris perfectly visible, cf. Figs. \ref{fig:samples_with_changes} and \ref{fig:samples_without_changes}. Contrary to previous claims found in the iris recognition literature, we did not find a single case with excessive pupil dilation.

Second, observing intra-class comparisons (violet lines in Figs. \ref{fig:CDF:IC}--\ref{fig:CDF:OS}) it is evident that dead irises can be encoded and recognized in more than 90\% of the cases, reaching even a perfect recognition for one method (IriCore). Since the first session was organized 5-7 hours after death, these results rebut the claim that the iris is useless soon after death (or minutes after death, as also is suggested in the literature).

Third, in the analysis of the following two cumulative distributions (blue and red lines in Figs. \ref{fig:CDF:IC}--\ref{fig:CDF:OS}), the dynamics of iris degradation can be observed. FNMR (false non-match rate) increases significantly to 48.88\% for the academic OSIRIS method, however, two commercial matchers -- IriTech and MIRLIN -- still present fairly good ability to recognize the samples (approx. 94.96\% and 82.76\% of correct verifications achieved for IriCore and MIRLIN methods, respectively). Note that these results are obtained for eye images captured approx. 16-21.5 hours after death. Surprisingly, we are still able to correctly recognize 73.33\% of irises 27 hours after death using IriCore method (cf. red graph in Fig. \ref{fig:CDF:IC}). Performance of the remaining methods is highly uneven, since they are able to recognize from 13.33\% (OSIRIS) to 60\% (VeriEye) of dead irises imaged 27 hours after dead. These results evidently show that the dynamics of post-mortem changes are much lower than commonly believed.

\subsection{Statistical significance of the observed results}

Although Figs. \ref{fig:CDF:IC}--\ref{fig:CDF:OS} suggest that distributions of genuine comparisons are distinctive, for completeness, a paired one-sided Kolmogorov-Smirnov test at the significance level $\alpha=0.05$ was performed to check whether the comparison scores calculated in our three experiments (for each matcher independently) come from the same distribution. The first null hypothesis is that scores between Session 2 samples and Session 1 samples (represented by blue graphs in Figs. \ref{fig:CDF:IC}--\ref{fig:CDF:OS}) come from the same distribution as the scores obtained in Session 1 (violet graphs in Figs. \ref{fig:CDF:IC}--\ref{fig:CDF:OS}). The alternative hypothesis is that the cumulative distribution of scores between Session 2 samples and Session 1 samples is larger (for VeriEye matcher) or smaller (for MIRLIN, OSIRIS and IriCore matchers) than cumulative distribution of scores obtained in Session 1. (Differences in alternative hypotheses definitions are due to different polarity of comparison scores across the matchers: MIRLIN, OSIRIS and IriCore generate dissimilarity scores, \ie, the higher the score, the worse the match, while VeriEye matcher generates similarity scores, \ie, the higher the score, the better the match). All tests (for all matchers) rejected the null hypotheses ($p$-value=0) in favor of the alternative hypotheses, hence the apparent differences in distributions observed in Figs. \ref{fig:CDF:IC}--\ref{fig:CDF:OS} are statistically significant. Statistical significance is also achieved when testing differences in scores between Session 3 samples and Session 1 samples (red graphs in Figs. \ref{fig:CDF:IC}--\ref{fig:CDF:OS}) vs scores obtained in Session 1 ($p$-value=0 for all tests formulated in an analogous way as above).





\section{Conclusions}
This paper presents the only study that we are aware of regarding post-mortem use of human iris as a biometric identifier. Contrary to claims common in the biometric community, our results show that human iris can be successfully employed for biometric authentication for a number of hours after death. Empirical study incorporating four different iris recognition method has shown that a significant portion of irises can be successfully recognized 5-7 hours after a person's demise (with FNMRs equaling 0\% to 8.33\% for the best and the worst performing method, respectively). However, this percentage is expected to decrease significantly as time period since death progresses, reaching FNMRs of 26.67\% to as much as 86.67\%. Comprehensive medical commentary is also presented to explain the underlying causes of such behavior, with processes assorted with corneal opacification, drying and loss of intraocular pressure recognized as the most probable sources of recognition errors.

The authors are aware of some limitations of this study that are inevitable to encounter while working simultaneously in the related, yet at the same time distant fields of biometrics, biology, pathology and medicine. In our future efforts we would like to focus on taking other possibly contributing factors into account, such as the influence of changing environmental conditions. We continue to expand the dataset, and we are working on necessary approvals that could enable us to release the data to the biometric community, to encourage more research in this important field.

\section{Acknowledgements}
The authors would like to thank Prof. Kevin Bowyer of the University of Notre Dame for his valuable comments on both the submission and the camera-ready version of this paper. This study had institutional review board clearance and the ethical principles of the Helsinki Declaration were followed by the authors.

{\small
\bibliographystyle{ieee}
\bibliography{refs}
}

\end{document}